\documentclass[letterpaper, 10 pt, conference]{ieeeconf}
\IEEEoverridecommandlockouts                       
\usepackage{cite}
\usepackage{amsmath,amssymb,amsfonts}
\usepackage{algorithm}
\usepackage{graphicx}
\usepackage{textcomp}
\usepackage{tabularx}
\usepackage[dvipsnames]{xcolor}
\usepackage{colortbl}
\usepackage{algpseudocode}
\usepackage{booktabs}
\usepackage{siunitx}
\usepackage{svg}
\usepackage{tikz}
\usepackage{amsmath,amsfonts}
\usepackage{dblfloatfix}
\renewcommand{\baselinestretch}{1}

\usepackage[caption=false,font=normalsize,labelfont=sf,textfont=sf]{subfig}
\usepackage{array,multirow,graphicx, calc}
\def\BibTeX{{\rm B\kern-.05em{\sc i\kern-.025em b}\kern-.08em
    T\kern-.1667em\lower.7ex\hbox{E}\kern-.125emX}}
\usepackage[colorlinks=true,allcolors=black]{hyperref}
\usepackage[nolist]{acronym} 
\usepackage{booktabs}  
\usepackage{float} 
\usepackage{multicol}  
\usepackage{mathtools}  
\let\labelindent\relax
\usepackage{enumitem}  
\usepackage{cite}
\usepackage[most]{tcolorbox}

\definecolor{Gray}{gray}{0.9}
\definecolor{LightGray}{HTML}{f4f6f6}
\definecolor{newText}{HTML}{159020}

\colorlet{shadecolor}{black!10} 
\makeatletter
\newcommand\fs@norules{\def\@fs@cfont{\bfseries}\let\@fs@capt\floatc@ruled
  \def\@fs@pre{}%
  \def\@fs@post{}%
  \def\@fs@mid{\kern3pt}%
  \let\@fs@iftopcapt\iftrue}
\makeatother
\floatstyle{norules}
 
\definecolor{cite_color}{RGB}{ 0, 0, 0 }
\definecolor{links_color}{RGB}{ 0, 0, 0 }

\newlength\myheight
\newlength\mydepth
\settototalheight\myheight{Xygp}
\def\BibTeX{{\rm B\kern-.05em{\sc i\kern-.025em b}\kern-.08em
    T\kern-.1667em\lower.7ex\hbox{E}\kern-.125emX}}

\newlength{\Oldarrayrulewidth}
\newcommand{\Cline}[2]{%
  \noalign{\global\setlength{\Oldarrayrulewidth}{\arrayrulewidth}}%
  \noalign{\global\setlength{\arrayrulewidth}{#1}}\cline{#2}%
  \noalign{\global\setlength{\arrayrulewidth}{\Oldarrayrulewidth}}}

\makeatletter
\def\blx@maxline{77}
\makeatother
\definecolor{block-gray}{gray}{0.9}
\newtcolorbox{myquote}{ arc=8pt, boxsep = 10pt,sidebyside,sidebyside align=top, rounded corners,colback=block-gray,grow to right by=-1mm,grow to left by=-1mm, boxrule=-1pt,boxsep=-1pt,breakable,left=5pt, right=5pt,top = 5pt, bottom= 5pt}
\makeatletter
\def\quoteparse{\@ifnextchar`{\quotex}{\singlequote}}
\def\quotex#1{\@ifnextchar`{\triplequote\@gobble}{\doublequote}}
\makeatother
\def\singlequote#1`{[StartQ]#1[EndQ]\quoteON}
\def\doublequote#1``{[StartQQ]#1[EndQQ]\quoteON}
\long\def\triplequote#1```{\begin{myquote}\parskip 1ex#1\end{myquote}\quoteON}
\def\quoteON{\catcode``=\active} 
\def\quoteOFF{\catcode``=12}
\quoteON
\def`{\quoteOFF \quoteparse}
\quoteOFF

\begin{document}

\newcommand{\STAB}[1]{\begin{tabular}{@{}c@{}}#1\end{tabular}}

\title{ \LARGE \bf Evaluating Robot Program Performance with Power Consumption–Driven Metrics in Lightweight Industrial Robots 
}

\author{ Juan Heredia$^{1}$, Emil Stubbe Kolvig-Raun$^{1}$, Sune Lundø Sørensen$^{1}$, and Mikkel Baun Kjærgaard$^{1}$ \thanks{ $^{1}$ Juan Heredia,  Emil Stubbe Kolvig-Raun, Sune Lundø Sørensen, and Mikkel Baun Kjægaard are with the Maersk Mc-Kinney Moller Institute, University of Southern Denmark, 5230 Odense, Denmark (e-mail: jehm@mmmi.sdu.dk; eskr@mmmi.sdu.dk; slso@mmmi.sdu.dk; mbkj@mmmi.sdu.dk). }
\thanks{This work is supported by the Innovation Fund Denmark for the project FERA (3149-00014A) }
}

\maketitle


\begin{abstract}

The code performance of industrial robots is typically analyzed through CPU metrics, which overlook the physical impact of code on robot behavior. This study introduces a novel framework for assessing robot program performance from an embodiment perspective by analyzing the robot’s electrical power profile. Our approach diverges from conventional CPU-based evaluations and instead leverages a suite of normalized metrics, namely, the energy utilization coefficient ($f_U$), the energy conversion metric ($f_C$), and the reliability coefficient ($f_R$), to capture how efficiently and reliably energy is used during task execution. Complementing these metrics, the established robot wear metric ($\alpha$) provides further insight into long-term reliability. Our approach is demonstrated through an experimental case study in machine tending, comparing four programs with diverse strategies using a UR5e robot. The proposed metrics directly compare and categorize different robot programs, regardless of the specific task, by linking code performance to its physical manifestation through power consumption patterns. Our results reveal the strengths and weaknesses of each strategy, offering actionable insights for optimizing robot programming practices. Enhancing energy efficiency and reliability through this embodiment-centric approach not only improves individual robot performance but also supports broader industrial objectives such as sustainable manufacturing and cost reduction.

\end{abstract}

\section{Introduction}

For over half a century, industrial robots have been essential in manufacturing. Recently, modern lightweight robots have transformed traditional industrial settings. These advanced robots, with built-in safety features, enable shared workspaces with humans. The International Federation of Robots (IFR) predicts a significant increase in their use \cite{international_federation_of_robotics_world_2022}, highlighting their growing impact on future industrial operations.

While the rise of lightweight industrial robots brings exciting opportunities, it also necessitates a comprehensive approach to performance evaluation. Traditional assessments have largely focused on either the mechanical capabilities, such as accuracy, repeatability, and speed \cite{kirschner_towards_2021, iso9283}, or on computational efficiency, measured through CPU utilization, memory consumption, and latency \cite{weisz_robobench_2016, biddulph_comparing_2018}. However, these approaches overlook a critical dimension: the embodiment of the robot. In other words, while code performance is often judged by its computational footprint, it is equally important to understand how the execution of this code impacts the robot’s physical behavior, including its energy consumption and mechanical wear.

\begin{figure}[t!] 
\centering
\includegraphics[width=.8\linewidth]{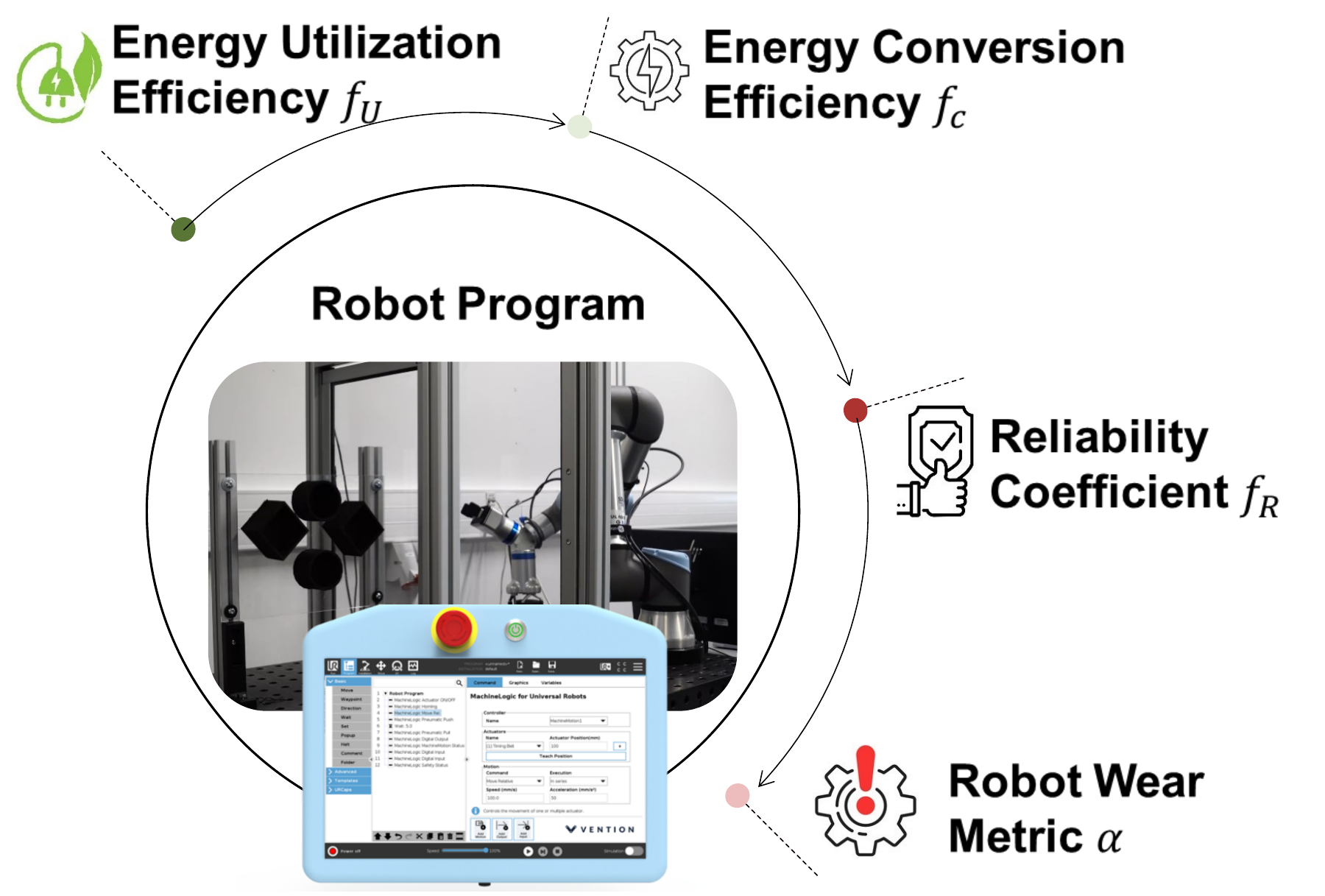}
\vspace{-.3cm} 
\caption{Assessing energy efficiency and reliability metrics in lightweight robots programs based on the robot power profile.}
\label{fig:Abstract}   
\vspace{-.7cm}
\end{figure}

This paper introduces a novel evaluation framework that assesses robot program performance from an embodiment perspective by analyzing the robot’s electrical power profile. Our approach quantifies the real-world impact in terms of energy efficiency and reliability of the robot program. Specifically, we introduce two normalized metrics for energy efficiency. The first metric, the energy utilization efficiency $f_U$, compares actual energy consumption to an idealized, loss-free model, measuring operational efficiency. The second metric, conversion efficiency $f_C$, assesses the conversion of electrical energy to mechanical work, indicating the proportion of energy effectively used. In addition to these energy metrics, we propose a Reliability Coefficient ($f_R$) based on the consistency of the robot’s power profile, which correlates power consumption patterns with performance stability. Complementing these, we employ an established robot wear metric ($\alpha$) \cite{10443045} to quantify the mechanical stress imposed on the robot’s joints over time.

We validate our integrated evaluation framework through a machine-tending case study using a UR5e robot, comparing four distinct programming strategies. This comprehensive analysis not only uncovers the strengths and weaknesses of each approach but also yields actionable insights for enhancing energy efficiency, reliability, and overall sustainability in industrial robotics. By linking code performance directly to physical outcomes, our work offers a deeper understanding of how programming choices affect energy dynamics and mechanical health, ultimately paving the way for more sustainable and reliable robotic operations in the future.

The structure of the paper is as follows: Section II provides an overview of the state-of-the-art from the literature. Sections III and IV present the energy efficiency metrics and reliability metrics in detail. In Section V, we showcase a case study applying the framework to a machine-tending scenario. Section VI discusses the findings and reflections from the case study, and finally, Section VII concludes the article.

\vspace{-1mm}
\section{Related Work}\label{sec:related}
\vspace{-1mm}

Assessing robot performance has long been a focal point in robotics research, particularly when it comes to energy efficiency and reliability metrics. Several studies have investigated absolute energy consumption as a key performance indicator in robotic manipulators \cite{heredia_data-driven_2021, Albonico21, paper20, vysocky_reduction_2020, Yuping23}. These works employ data-driven models to analyze power usage under varying operational conditions and to optimize robot operations. While they have significantly advanced our understanding of energy consumption across different tasks, a notable limitation is the absence of normalized metrics that account for variations in task complexity and execution time.

To address these gaps, researchers have proposed alternative metrics. For instance, metrics such as energy per distance traveled—or the “cost of transportation”—normalize energy usage by the work performed \cite{larsen_energy_2011}. In addition, a benchmarking framework based on a set of metrics that compares the energy performance of various lightweight robots from different manufacturers was presented in \cite{heredia_labelling_2023}. Other works have introduced metrics focusing on computational efficiency in robotic software, such as “$J/m$,” “$J/hour$” \cite{swanborn_energy_2020}, performance-per-watt (“$FPS^4/W$”) for evaluating processing efficiency on different hardware platforms \cite{cheng_performance_2018}, and the Energy Delay Product ($Et^n$) family \cite{Roberts17}. While these approaches provide valuable insights, they focus either on mechanical or computational efficiency in isolation, limiting their applicability to holistic robot performance assessment.

In parallel, research on robot reliability has traditionally relied on failure prediction methods such as fault-tree analysis, Markov models, and predictive maintenance techniques \cite{dhillon2015}. Recent studies, however, suggest that energy consumption profiles can serve as indicators of overall robot reliability and degradation \cite{heredia_data-driven_2021, heredia_ecdp_2023}. This shift toward using power data for reliability estimation aligns with our approach, which introduces a reliability coefficient based on energy consistency. Additionally, \cite{10443045} introduced a predictive model for estimating a robot’s end-of-life (EoL) using the stress metric ($\alpha$). While our method primarily focuses on runtime diagnostics and not EoL estimation, we recognize the value of this metric as an additional parameter. Incorporating it into our analysis allows for a more comprehensive assessment of the robot's performance and reliability.

Our research builds upon these foundations by proposing a comprehensive performance assessment framework that integrates energy efficiency, work performed, and reliability into a unified set of normalized metrics. Unlike previous studies that focus on single-dimensional measures, our approach enables fair comparisons between robotic programs regardless of task type. This directly supports the objectives outlined in the abstract: improving benchmarking, optimizing programming strategies, and enhancing energy-efficient robotics development. By addressing the limitations of existing methodologies, our proposed framework contributes to a more holistic evaluation of robot performance, paving the way for standardized sustainability metrics in industrial robotics.

\vspace{-2mm}
\section{Energy Efficiency Metrics}\label{sec:ee_metrics}

This section presents the theoretical foundations for our energy efficiency metrics, building on the energy disaggregation model presented in \cite{heredia_ecdp_2023}. Our approach leverages repeated experiments, conducted $n$ times, to capture detailed power consumption data during robot program execution. Only successful runs are considered when computing the mean value of each energy performance metric.

\subsection{\textbf{Energy Utilization Coefficient ($f_U$)}}


The energy utilization coefficient quantifies energy losses during task execution by comparing the ideal (basal) energy required with the actual energy consumed. It is defined as: 

\begin{equation}
    f_U = \frac{E_{B} }{E_{R} } \, ,
     \label{eq.fU}
\end{equation}
where $E_{B}$ and $E_{R}$ are the basal and the actual energy consumption during the experiment. 

{\begingroup
\footnotesize
\quoteON
``` \noindent \textit{\textbf{Note:} If ($f_U = 1$), the system behaves in ideal conditions (no losses). But, in a real-world scenario, the coefficient would be less than 1 ($f_U < 1$) due to real-world constraints.}```
\quoteOFF
\normalsize
\endgroup}


\subsubsection{Basal Energy Consumption ($E_B$)} This factor denotes the minimum energy needed for fundamental operations. In an ideal scenario, the robot program operates without energy losses from friction, heat dissipation, or other inefficiencies. $E_{B}$ includes several key components:


\begin{description}[style=unboxed, leftmargin=0.3cm]
     \item[Electronics Energy ($E_E$):]  Power consumed by the control system, sensors, and communication modules, even during idle periods or in standby mode. The electronics energy is calculated based on the electronics power, $E_E\, =\, P_E \, T$, following the protocol by \cite{heredia_ecdp_2023}, where $P_E$ is the electrical power of the electronics, and, $T$ is the duration of the task.
    \item[Mechanical Brakes Energy ($E_{MB}$):] Most of the lightweight robots utilize brakes to react in the case of a collision. These brakes consume a constant amount of power ($P_{MB}$) to keep released while the robot is in motion, such that ($E_{MB}\, =\, P_{MB} \, T$). 
    \item[\textbf {Mechanical work}:]  Under ideal conditions, mechanical work is fully conserved. The mechanical energy is equal to the change in internal energy of a system, such as $W = \Delta (E_P +E_K)$, where $E_P$ and $E_K$ are the potential and kinetic energy, respectively. 
\end{description}

The total basal energy can be expressed as: 
\begin{equation}
    E_B = (P_E + P_{MB}) \cdot T + \Delta E_P + \Delta E_K \, .
    \label{eq.ebasal}
\end{equation}

\subsubsection{Actual Energy Consumption ($E_R$)} represents the total energy used during the robot’s operation, including motors, controllers, sensors, and other components. $E_{R}$ is measured using an external power analyzer, recording the instantaneous power ($P_R$). The actual energy consumption, $E_R$, is then estimated by integrating the power over the operation time:

\begin{equation}
    E_{R} = \int P_R(t)\, dt \, .
     \label{eq.ER}
     \vspace{-3mm}
\end{equation}


\subsection{\textbf{Energy Conversion Coefficient ($f_C$)}}

The energy conversion coefficient measures how effectively a robotic system converts electrical energy into the mechanical work needed to accomplish a task. In our approach, we focus exclusively on the positive values of mechanical work. This focus is intentional because our primary interest lies in quantifying the energy required to generate mechanical motion. Negative mechanical power, often associated with regenerative modes, is not considered since most industrial robots lack batteries for energy storage, and any energy recovered is typically dissipated through resistors as heat. Accordingly, the energy conversion coefficient is defined as: 
\begin{equation}
    f_C = \frac{E_{MP} }{E_{R}} \, ,
     \label{eq.fC}
\end{equation}
where $E_{MP}$ represents the total positive mechanical energy generated, and $E_{R}$ is the total actual energy consumption.
  
{\begingroup
\quoteON
\footnotesize
```\noindent \textit{\textbf{Note:} \(f_C = 1\) signifies perfect energy conversion, while in reality \(f_C < 1\) because the robot requires energy to energize other components as well to compensate for electrical and mechanical losses.}```
\normalsize
\quoteOFF
\endgroup}

\subsubsection{Mechanical Output Energy ($E_M$)} is derived from the work performed by each joint. For the i-th joint, the instantaneous mechanical power is given by:
\begin{equation}
    P_{Mi} = \tau_i \Dot{\theta_i} \,,
    \vspace{-1mm}
\end{equation}
where $P_M$ represents the mechanical power output, $\tau_i$ denotes the torque exerted, and $\dot{\theta_i}$ represents the angular velocity for the given i-th joint.

The mechanical power of a motor operates in two distinct regions: consumption (positive mechanical power) and generation (negative mechanical power). When the regenerated energy is excessive, it is often dissipated through resistive elements. To focus on the energy that contributes to mechanical work, we consider only the positive mechanical power:
\vspace{-2mm}
\begin{equation}
    P_{Mi}^+ = 
    \begin{cases}
     \text{$\tau_i (t) \, \dot{\theta}_i (t)$}  &  \text{if $ \tau_i(t)  \,  \dot{\theta}_i(t) > 0$ } \,  \\
      0 &  \text{if $ \tau_i(t)  \, \dot{\theta}_i(t) <0$}   \, 
    \end{cases} \, .
\end{equation}

Then, the positive mechanical energy ($E_{MP}$) could be estimated by:
\vspace{-2mm}
\begin{equation}
    E_{MP} =  \sum_{i=1}^{n_J} \int_0^T P_{Mi}^+ dt \, ,
     \label{eq.EM}
\end{equation}
where $n_J$ is the number of joints and $T$ is the total duration of the task.

This formulation of ($f_C$) ensures that the metric reflects the energy exclusively dedicated to performing the task, without the confounding effects of regenerative energy, which in many cases is not recoverable due to hardware limitations.

\section{Reliability Metrics}\label{sec:r_metrics}

This section introduces a novel approach to evaluate robotic reliability using statistical metrics derived from power consumption profiles. Traditionally, reliability, defined as the probability of a robot performing its function under given conditions, has been assessed through measures such as positional accuracy and repeatability \cite{zhang_positioning_2021} using techniques like Markov models or fault-tree analysis \cite{dhillon2015}. In contrast, our method leverages the consistency of power consumption to infer performance uniformity and overall system health. This approach provides insight into maintenance needs and helps optimize performance.

The reliability coefficient ($f_R$) is quantified by combining three reliability metrics ($c_1$, $c_2$, and $c_3$). In addition, the robot wear metric ($\alpha$) is employed to capture the stress induced by the control program \cite{10443045}. We also note that sensitivity analysis of these parameters is recommended to assess robustness against measurement uncertainties.

\subsection{\textbf{Reliability Coefficient} ($f_R$)}
The reliability coefficient ($f_R$) aggregates multiple aspects of power profile similarity across $n$ tests:
\begin{equation}
    f_R = \frac{w_1 \, c_1 + w_2 \, c_2 + w_3 \, |c_3|}{w_1+w_2+w_3} \, ,
     \label{eq.fR}
\end{equation}
where $w_1$, $w_2$, and $w_3$ are weights assigned to each individual metric. 

\quoteON
\footnotesize
```\noindent \textit{\textbf{Note:} $f_R$ ranges between 0 and 1. Values near 1 indicate highly consistent power profiles, suggesting a reliable program, whereas lower values indicate variability and reduced reliability.}```
\normalsize
\quoteOFF

\subsubsection{Root Mean Squared Error Coefficient ($c_1$)} measures the deviation between an individual power profile ($P_i(t)$) and the mean profile of successful runs ($\overline{P_S}(t)$). The mean profile $\overline{P_S}$ is defined as:  
\begin{equation}
    \overline{P_S}(t) = \frac{1}{n_{succ}} \sum_i^n c_{succ} \, P_i(t) \, ,
    \vspace{-2mm}
\end{equation}
where $n$, $n_{succ}$, $c_{succ}$ represent number of repetitions, the number of successful samples, and the success coefficient. The success coefficient is 1 when the robot completes the task and 0 when it fails. 

\begin{figure*}[t!]
\centering

\includegraphics[width=.9\linewidth]{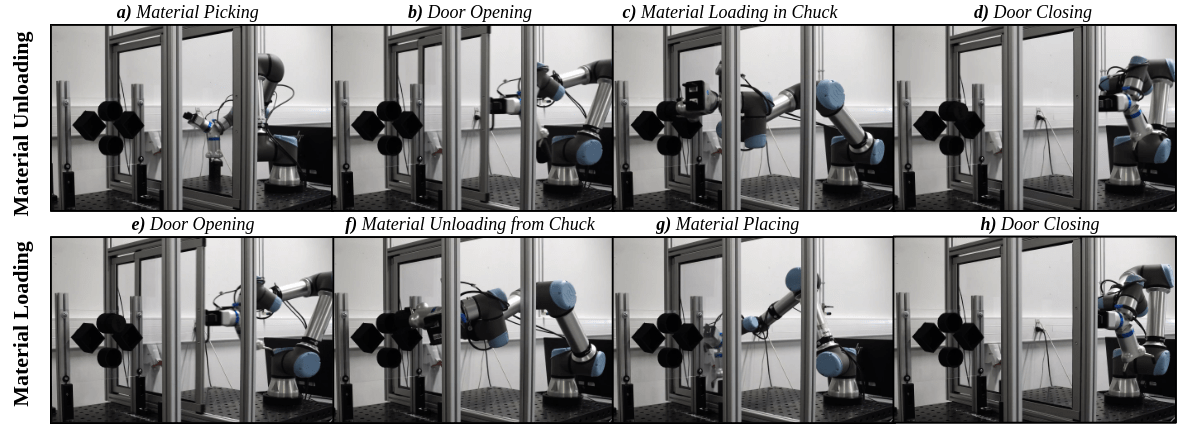}
\vspace{-3mm}
\caption{The top four figures show the loading process of the task, where the robot picks the object from the ``Picking Position," opens the mockup door, loads the object onto the mockup chuck, and closes the mockup door. The bottom four figures present the unloading process, where the robot opens the mockup door, unloads the object from the mockup chuck, places the object in the ``Unloading Position," and closes the mockup door.}
\label{fig:CNC}
\vspace{-.6cm}
\end{figure*}

Then, we compare each power profile from an individual experiment to the mean power of the successful samples. This approach eliminates the possibility of considering programs as reliable when the robot consistently and repeatedly fails the task. The metric normalized root mean squared error ($\epsilon$) is calculated by employing the following equation:
\begin{equation}
    \epsilon = \frac{1}{R(P_i)} \sqrt{ \frac{1}{T} \int_0^T (P_i(t) - \overline{P_S}(t))^2 dt} \, ,
\end{equation}
where $R(P_i)$ is defined as the difference between its maximum and minimum values ($R(P) = P_{\text{max}} - P_{\text{min}}$).

Finally, the first metric is defined, such as:
\vspace{-1mm}
\begin{equation}
    c_1 = 1 - \epsilon \, .
     \label{eq.c1}
\end{equation} 
\vspace{-4mm}
\quoteON
\footnotesize
```\noindent \textit{\textbf{Note:} If ($c_1 = 1$), the power profiles of each of the samples are similar to each other. However, in unreliable programs, this coefficient would be less than 1 ($c_1 < 1 $), indicating that the power profiles of these programs vary from one another.}```
\normalsize
\quoteOFF

\subsubsection{Coefficient of Variation ($c_2$)} quantifies the relative variability of the power consumption signal at each timestamp. In this context, it is equivalent to the normalized root-mean-square deviation (NRMSD) or percent RMS. For each timestamp, the power measurements across repeated experiments form an independent distribution. The instantaneous coefficient of variation is defined as the ratio of the standard deviation $\sigma$(t) to the mean power consumption $\overline{P_S}(t)$ at that timestamp, such that:
\begin{equation}
    cv(t) = \frac{\sigma (t) } {\overline{P_S}(t)} \, ,
    \vspace{-1mm}
\end{equation}

To capture the overall variability throughout the entire task duration $T$, we compute the time-averaged coefficient of variation. We then define the metric $c_2$ as follows:
\begin{equation}
    c_2 = 1 - \frac{1}{T} \int_0^T cv(t) \, dt \, .
     \label{eq.c2}
\end{equation}

\quoteON
\footnotesize
```\noindent \textit{\textbf{Note:} A \(c_2\) value of 1 indicates minimal variation in the power profiles, while lower values imply higher variability and lower reliability.}```
\normalsize
\quoteOFF

\subsubsection{Cross Correlation Coefficient ($c_3$)} is used to assess the similarity between the power consumption profile of an individual run and the mean power profile derived from successful trials. We employ Pearson’s linear correlation coefficient ($\rho$) as the basis for this metric. Pearson’s coefficient is defined as:
\begin{equation}
 \rho = \frac{\text{cov}(P_X, P_Y)}{\sigma_X \sigma_Y} \, ,    
\end{equation}
where $\text{cov}$ denotes the covariance between signals $P_X$ and $P_Y$, while $\sigma_X$ and $\sigma_Y$ represent the standard deviation of  $P_X$ and $P_Y$, respectively.

To compute this metric, each signal is paired with the mean power consumption profile of successful cases, such that:
\vspace{-1mm}
\begin{equation}
 c_3 = \frac{\text{cov}(P_i, \overline{P_S})}{\sigma_{P_i} \sigma_{\overline{P_S}}} \, .
  \label{eq.c3}
\end{equation}

\footnotesize
\quoteON
```\noindent \textit{\textbf{Note:} The correlation coefficient is normalized and ranges between -1 and +1. A coefficient of -1 signifies one signal is the exact opposite of the other. Conversely, a coefficient of +1 denotes a perfect positive correlation, indicating identical ordering between the two signals. A coefficient of 0 means no relationship exists between the signals being compared.}```
\quoteOFF
\normalsize

\subsection{\textbf{Robot wear metric} ($\alpha$)}



The  $\alpha$ metric proposed by \cite{10443045} quantifies the stress or wear induced on the robot by the control program. It measures stress for each joint by considering the Joint Signal Fluxes (JSFs), which are the discrepancies between target and actual currents during operation.

The JSF is adjusted by a weighting function, $\omega$, accounting for friction loss due to rotational velocity and temperature. It is standardized using calibration constants ($\phi$, $\tau_k$, and $\tau_{max}$) specific to the gear type. Consequently,  $\alpha$ represents a time series of stress measurements for each joint during program execution. This metric allows for comparison across different joint sizes and robot models, as depicted in Eq. \ref{eq:alpha}:
\begin{equation}
  \label{eq:alpha}
  \alpha_{j} = \{f(\epsilon,\omega, \phi, \tau_{k}, \tau_{max})\}_{j} \, ,
\end{equation}
where $j$ denotes a joint index (0 to 5), $\epsilon$ the JSF, $\omega$ the weighing function, $\phi$ the gear ratio (number of teeth), $\tau_{k}$ a torque constant, and $\tau_{max}$ the maximum torque that the gear can endure.

In assessing program performance with respect to induced stress, we employ a method that involves calculating the average of the infimums across each sequence of $\alpha$ values, denoted $\alpha_{S}$. These averages are then accumulated for each joint, as depicted in Eq. \ref{eq:eval_alpha},
\begin{equation}
  \label{eq:eval_alpha}
  \alpha_{S} = \frac{1}{n} \sum_{i=0}^{n-1} \inf(\alpha_{j}) \, , 
\end{equation}
where $n$ is the number of repetitions. The infimums, $inf(\alpha_j)$, are determined for each joint by sorting the $\alpha$ sequence in descending order and taking its maximum value.

\quoteON
\footnotesize
```\noindent \textit{\textbf{Note:} The greater the $\alpha_{S}$ value, the greater the stress. A value equal to 0 signifies that the robot is stress-free.}```
\normalsize
\quoteOFF

\begin{table*}[b!]

\vspace{-.4cm}

\centering
\caption{Energy efficiency and reliability metrics of four programs with two different waiting times (machine operation).}

\vspace{-.4cm}

\resizebox{1\textwidth}{!}{%
\addtolength{\tabcolsep}{-0.2em}   
\def\arraystretch{1.6}%
\begin{tabular}{>{\cellcolor{white}}p{\dimexpr.02\linewidth-2\tabcolsep-1.3333\arrayrulewidth}p{\dimexpr.025\linewidth-2\tabcolsep-1.3333\arrayrulewidth}ccccccccccc}
\midrule\toprule[-2mm]
\cellcolor{white} & & \multicolumn{5}{c}{\textit{{Energy Efficiency Metrics}}}  & & \multicolumn{5}{c}{\textit{{Reliability Metrics}}} \\ 
\Cline{0.5pt}{3-7} \Cline{0.5pt}{9-13} 
  \\[-4.9mm]

\cellcolor{white} & P. & $\overline{E}_B$ (J)  &  $\overline{E}_R$ (J) &  $f_U$  &  $\overline{E}_{MP}$ (J) &  $f_C$ & & $c_1$ & $c_2$& $c_3$ & $f_R$ & $\alpha_S$ \\
\rowcolor{LightGray}
 & \textit{A} & \begin{minipage}{.075\linewidth} 
  \includegraphics[trim={0 0 50 0},clip,width=\linewidth]{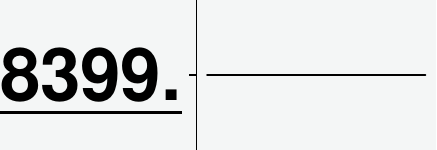}
\end{minipage}  & \begin{minipage}{.095\linewidth} 
  \includegraphics[width=\linewidth]{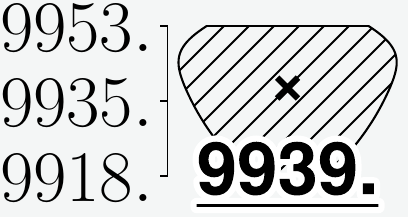}
\end{minipage} & \begin{minipage}{.095\linewidth} 
  \includegraphics[width=\linewidth]{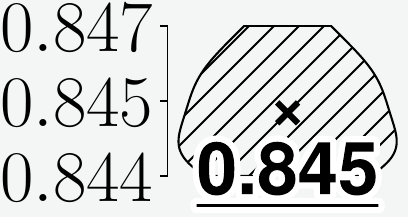}
\end{minipage} & \begin{minipage}{.095\linewidth}  
  \includegraphics[width=\linewidth]{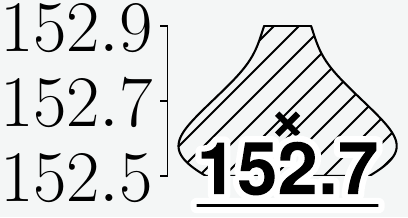}
\end{minipage} & \begin{minipage}{.095\linewidth} 
  \includegraphics[width=\linewidth]{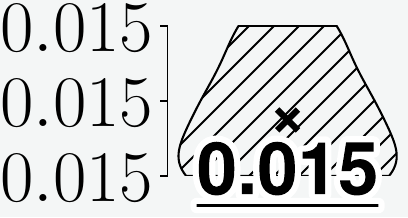}
\end{minipage} & & \begin{minipage}{.095\linewidth}
  \includegraphics[width=\linewidth]{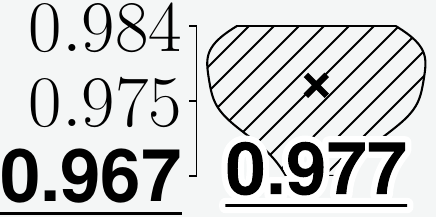}
\end{minipage} & \begin{minipage}{.095\linewidth}
  \includegraphics[width=\linewidth]{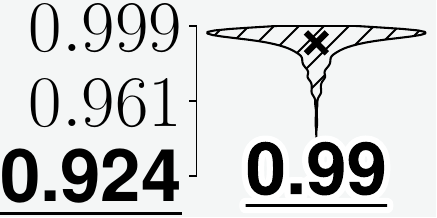}
\end{minipage} & \begin{minipage}{.095\linewidth}
  \includegraphics[width=\linewidth]{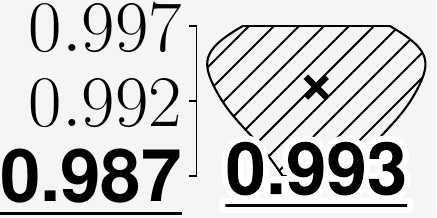}
\end{minipage} & \begin{minipage}{.095\linewidth}
  \includegraphics[width=\linewidth]{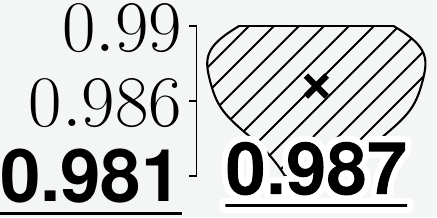}
\end{minipage} & \begin{minipage}{.095\linewidth} 
  \includegraphics[width=\linewidth]{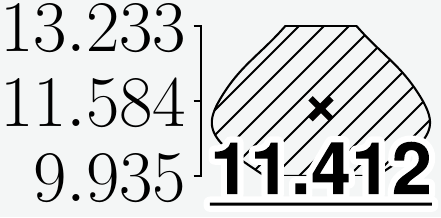}
\end{minipage} \\
& \textit{B} & \begin{minipage}{.075\linewidth} 
  \includegraphics[trim={0 0 50 0},clip,width=\linewidth]{results/ProgramA_10s_E_B.pdf}
\end{minipage} & \begin{minipage}{.095\linewidth} 
  \includegraphics[width=\linewidth]{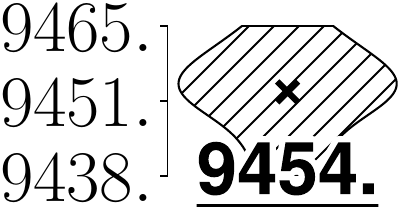}
\end{minipage} & \begin{minipage}{.095\linewidth}
  \includegraphics[width=\linewidth]{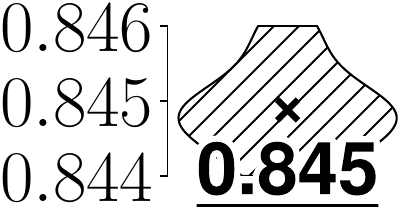}
\end{minipage} & \begin{minipage}{.095\linewidth}  
  \includegraphics[width=\linewidth]{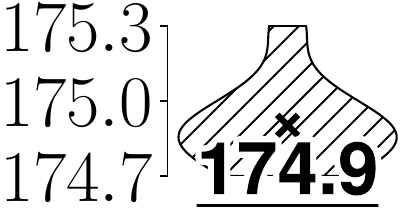}
\end{minipage} & \begin{minipage}{.095\linewidth}
  \includegraphics[width=\linewidth]{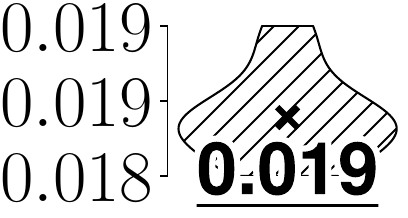}
\end{minipage} & & \begin{minipage}{.095\linewidth}
  \includegraphics[width=\linewidth]{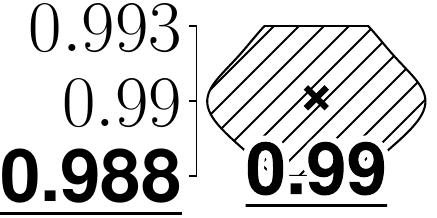}
\end{minipage} & \begin{minipage}{.095\linewidth}
  \includegraphics[width=\linewidth]{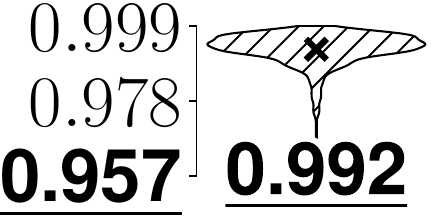}
\end{minipage} & \begin{minipage}{.095\linewidth}
  \includegraphics[width=\linewidth]{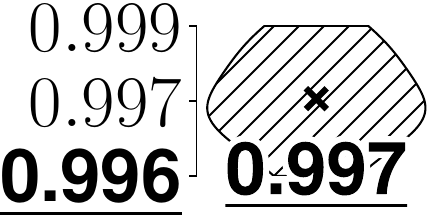}
\end{minipage} & \begin{minipage}{.095\linewidth}
  \includegraphics[width=\linewidth]{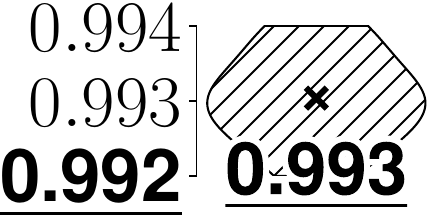}
\end{minipage} & \begin{minipage}{.095\linewidth} 
  \includegraphics[width=\linewidth]{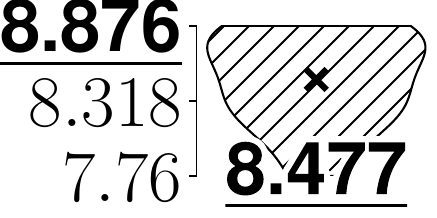}
\end{minipage} \\
\rowcolor{LightGray}
& \textit{C} &  \begin{minipage}{.075\linewidth} 
  \includegraphics[trim={0 0 50 0},clip,width=\linewidth]{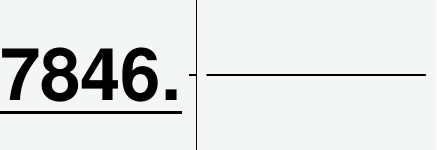}
\end{minipage} & \begin{minipage}{.095\linewidth} 
  \includegraphics[width=\linewidth]{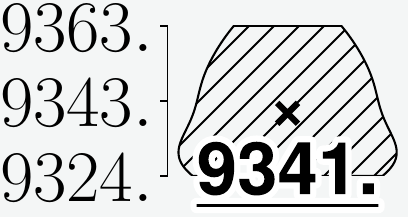}
\end{minipage} & \begin{minipage}{.095\linewidth}
  \includegraphics[width=\linewidth]{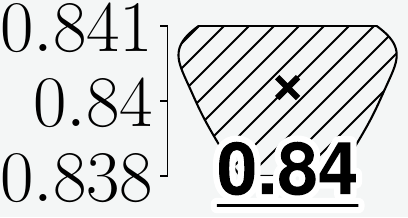}
\end{minipage} & \begin{minipage}{.095\linewidth}  
  \includegraphics[width=\linewidth]{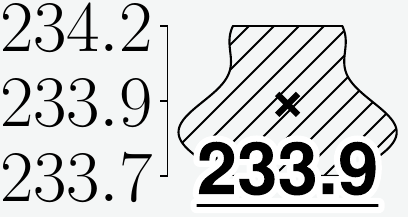}
\end{minipage} & \begin{minipage}{.095\linewidth}
  \includegraphics[width=\linewidth]{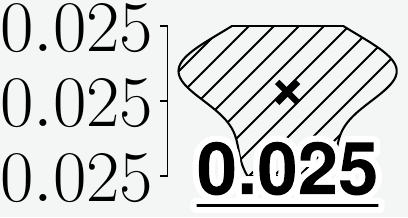}
\end{minipage} & & \begin{minipage}{.095\linewidth}
  \includegraphics[width=\linewidth]{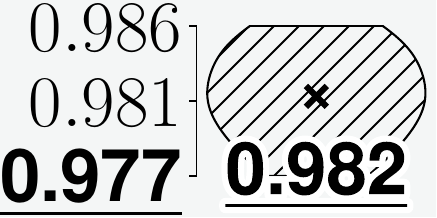}
\end{minipage} & \begin{minipage}{.095\linewidth}
  \includegraphics[width=\linewidth]{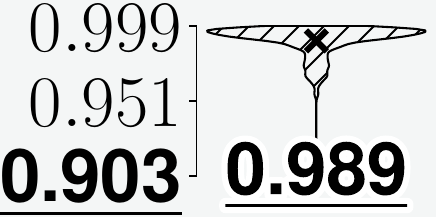}
\end{minipage} & \begin{minipage}{.095\linewidth}
  \includegraphics[width=\linewidth]{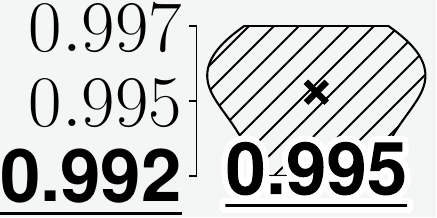}
\end{minipage} &\begin{minipage}{.095\linewidth}
  \includegraphics[width=\linewidth]{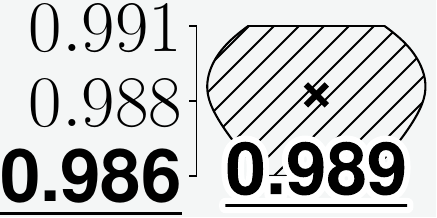}
\end{minipage} & \begin{minipage}{.095\linewidth} 
  \includegraphics[width=\linewidth]{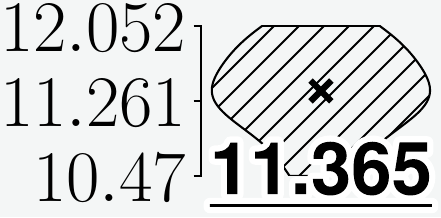}
\end{minipage}\\ 
\multirow{-7}{*}{\rotatebox[origin=c]{90}{Wait time: \textbf{10 Seconds}}}  
& \textit{D} & \begin{minipage}{.075\linewidth}  
  \includegraphics[trim={0 0 50 0},clip,width=\linewidth]{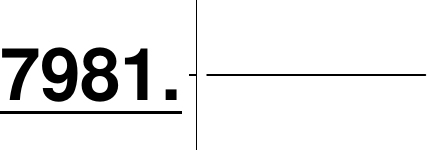}
\end{minipage}  & \begin{minipage}{.095\linewidth}  
  \includegraphics[width=\linewidth]{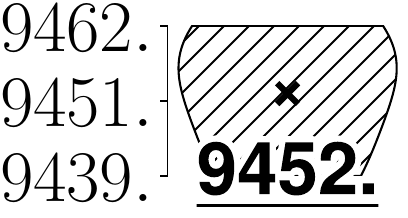}
\end{minipage} & \begin{minipage}{.095\linewidth}
  \includegraphics[width=\linewidth]{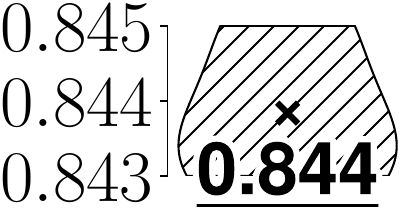} 
\end{minipage} & \begin{minipage}{.095\linewidth}   
  \includegraphics[width=\linewidth]{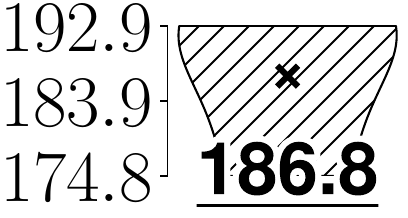}
\end{minipage} & \begin{minipage}{.095\linewidth}
  \includegraphics[width=\linewidth]{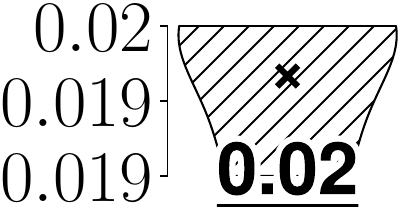}
\end{minipage} & & \begin{minipage}{.095\linewidth}
  \includegraphics[width=\linewidth]{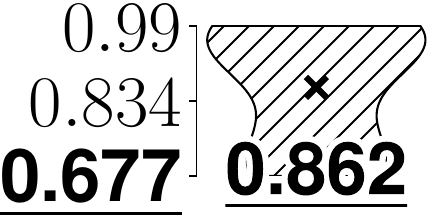}
\end{minipage} & \begin{minipage}{.095\linewidth} 
  \includegraphics[width=\linewidth]{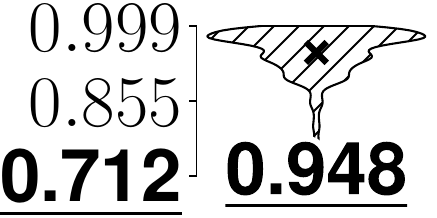}
\end{minipage} & \begin{minipage}{.095\linewidth} 
  \includegraphics[width=\linewidth]{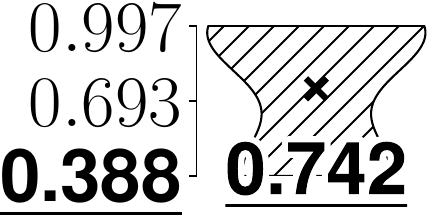}
\end{minipage} & \begin{minipage}{.095\linewidth}
  \includegraphics[width=\linewidth]{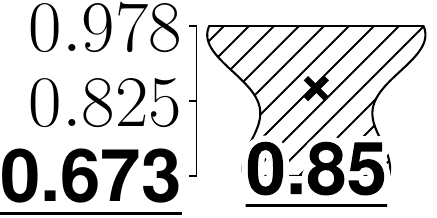}
\end{minipage} & \begin{minipage}{.095\linewidth} 
  \includegraphics[width=\linewidth]{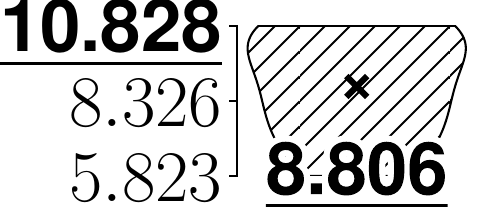}
\end{minipage} \\\midrule
\rowcolor{LightGray}
 & \textit{A} & \begin{minipage}{.075\linewidth} 
  \includegraphics[trim={0 0 50 0},clip,width=\linewidth]{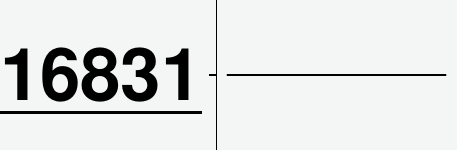}
\end{minipage}  & \begin{minipage}{.095\linewidth} 
  \includegraphics[width=\linewidth]{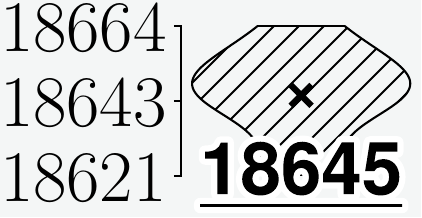}
\end{minipage} & \begin{minipage}{.095\linewidth} 
  \includegraphics[width=\linewidth]{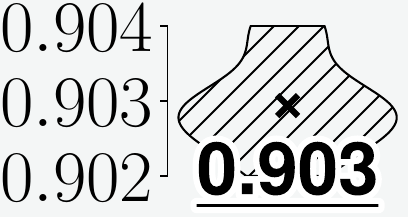}
\end{minipage} & \begin{minipage}{.095\linewidth}  
  \includegraphics[width=\linewidth]{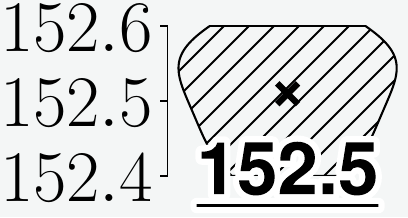}
\end{minipage} & \begin{minipage}{.095\linewidth}  
  \includegraphics[width=\linewidth]{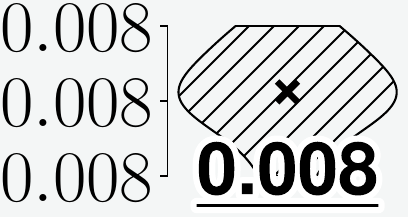}
\end{minipage} & & \begin{minipage}{.095\linewidth}
  \includegraphics[width=\linewidth]{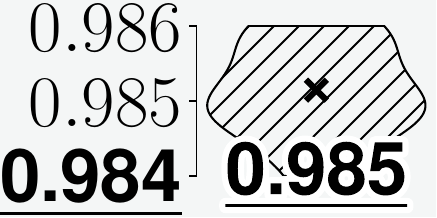}
\end{minipage} & \begin{minipage}{.095\linewidth}
  \includegraphics[width=\linewidth]{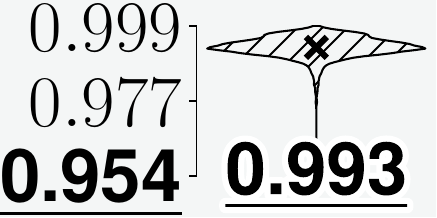}
\end{minipage} & \begin{minipage}{.095\linewidth}
  \includegraphics[width=\linewidth]{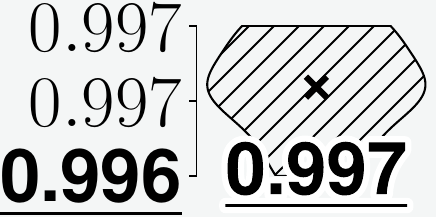}
\end{minipage} & \begin{minipage}{.095\linewidth}
  \includegraphics[width=\linewidth]{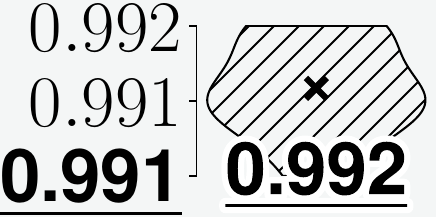}
\end{minipage} & \begin{minipage}{.095\linewidth} 
  \includegraphics[width=\linewidth]{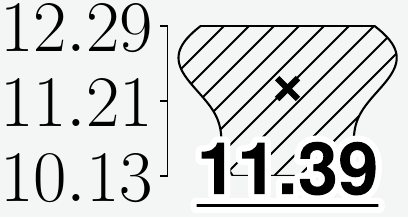}
\end{minipage}\\
& \textit{B} & \begin{minipage}{.075\linewidth} 
  \includegraphics[trim={0 0 50 0},clip,width=\linewidth]{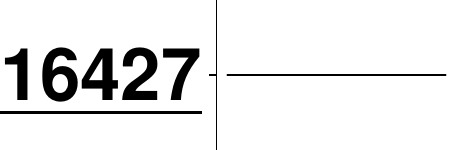}
\end{minipage} & \begin{minipage}{.095\linewidth} 
  \includegraphics[width=\linewidth]{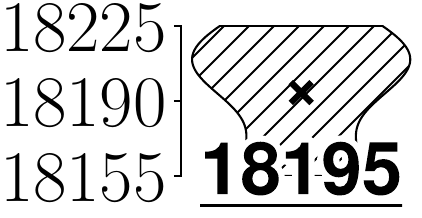}
\end{minipage} & \begin{minipage}{.095\linewidth}
  \includegraphics[width=\linewidth]{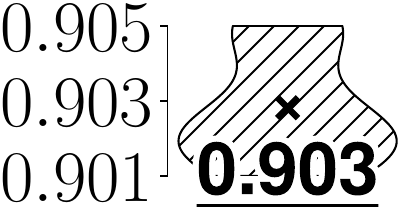}
\end{minipage} & \begin{minipage}{.095\linewidth}  
  \includegraphics[width=\linewidth]{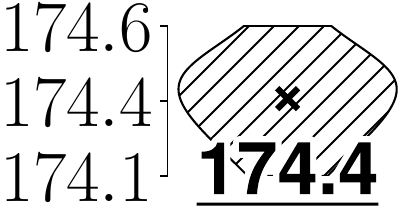}
\end{minipage} & \begin{minipage}{.095\linewidth}
  \includegraphics[width=\linewidth]{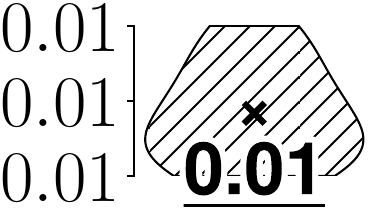}
\end{minipage} & & \begin{minipage}{.095\linewidth}
  \includegraphics[width=\linewidth]{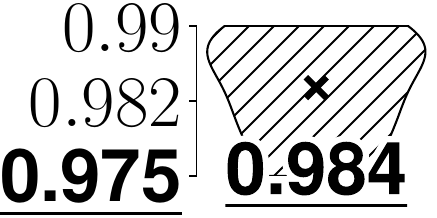}
\end{minipage} & \begin{minipage}{.095\linewidth}
  \includegraphics[width=\linewidth]{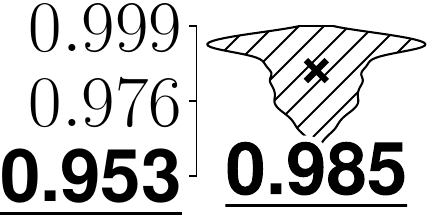}
\end{minipage} & \begin{minipage}{.095\linewidth}
  \includegraphics[width=\linewidth]{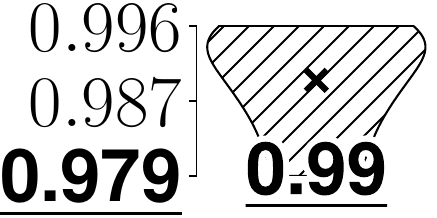}
\end{minipage} & \begin{minipage}{.095\linewidth}
  \includegraphics[width=\linewidth]{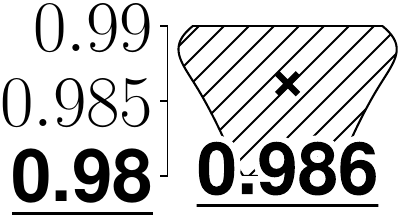}
\end{minipage} & \begin{minipage}{.095\linewidth} 
  \includegraphics[width=\linewidth]{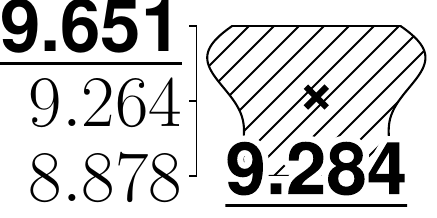}
\end{minipage} \\

\rowcolor{LightGray}
& \textit{C} & \begin{minipage}{.075\linewidth} 
  \includegraphics[trim={0 0 50 0},clip,width=\linewidth]{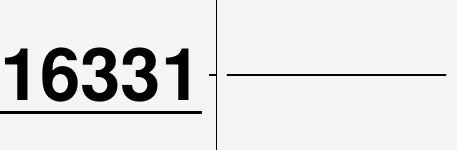}
\end{minipage}  & \begin{minipage}{.095\linewidth} 
  \includegraphics[width=\linewidth]{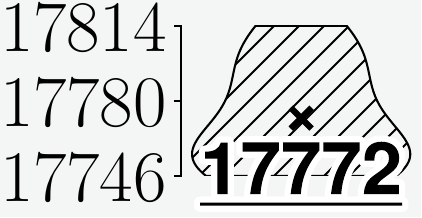}
\end{minipage} & \begin{minipage}{.095\linewidth}
  \includegraphics[width=\linewidth]{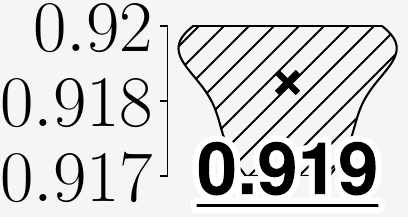}
\end{minipage} & \begin{minipage}{.095\linewidth}  
  \includegraphics[width=\linewidth]{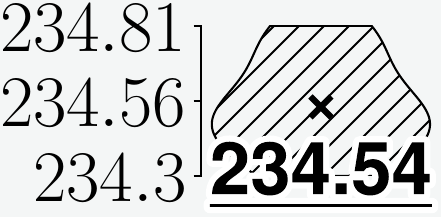}
\end{minipage} & \begin{minipage}{.095\linewidth}
  \includegraphics[width=\linewidth]{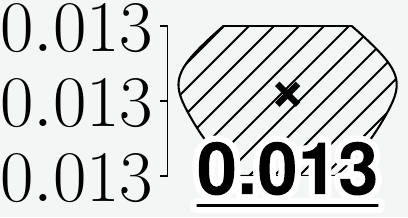}
\end{minipage} & & \begin{minipage}{.095\linewidth} 
  \includegraphics[width=\linewidth]{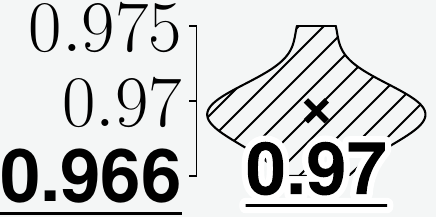}
\end{minipage} & \begin{minipage}{.095\linewidth}
  \includegraphics[width=\linewidth]{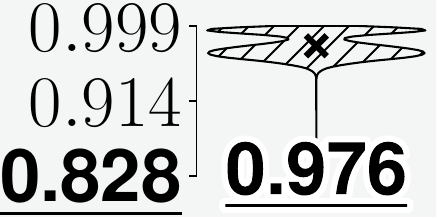}
\end{minipage} & \begin{minipage}{.095\linewidth}
  \includegraphics[width=\linewidth]{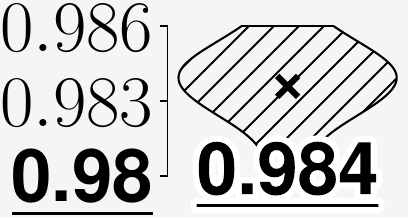}
\end{minipage} &\begin{minipage}{.095\linewidth}
  \includegraphics[width=\linewidth]{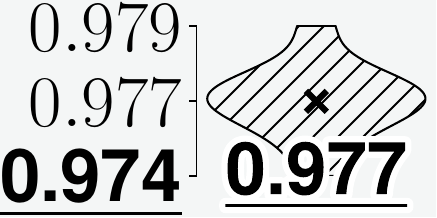}
\end{minipage} & \begin{minipage}{.095\linewidth} 
  \includegraphics[width=\linewidth]{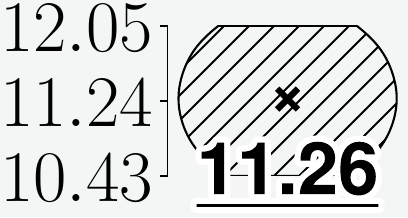}
\end{minipage}\\
\multirow{-7}{*}{\rotatebox[origin=c]{90}{Wait time: \textbf{100 Seconds}}} 
& \textit{D} & \begin{minipage}{.075\linewidth} 
  \includegraphics[trim={0 0 50 0},clip,width=\linewidth]{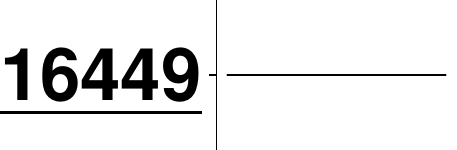}
\end{minipage} & \begin{minipage}{.095\linewidth} 
  \includegraphics[width=\linewidth]{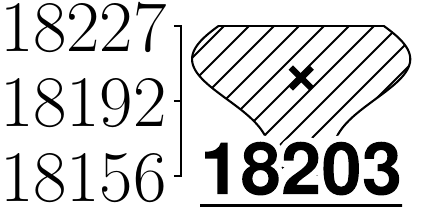}
\end{minipage} & \begin{minipage}{.095\linewidth}
  \includegraphics[width=\linewidth]{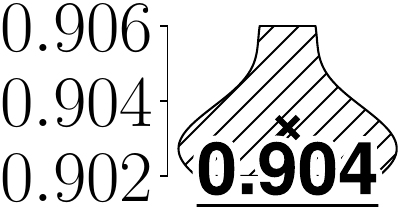}
\end{minipage} & \begin{minipage}{.095\linewidth}   
  \includegraphics[width=\linewidth]{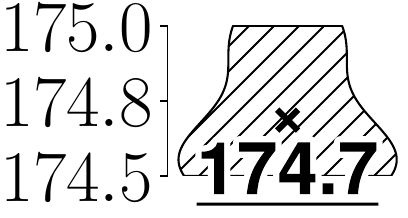}  
\end{minipage} & \begin{minipage}{.095\linewidth}
  \includegraphics[width=\linewidth]{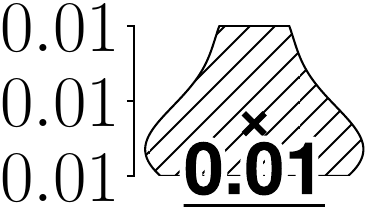}
\end{minipage} & & \begin{minipage}{.095\linewidth}
  \includegraphics[width=\linewidth]{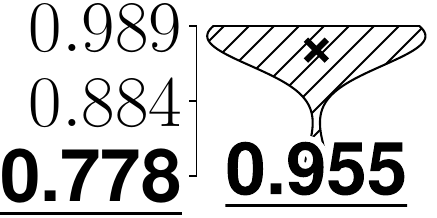} 
\end{minipage} & \begin{minipage}{.095\linewidth}
  \includegraphics[width=\linewidth]{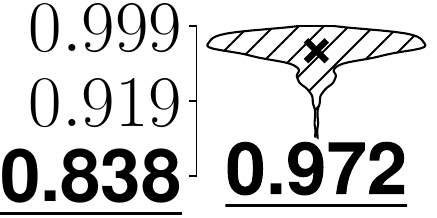}
\end{minipage} & \begin{minipage}{.095\linewidth}
  \includegraphics[width=\linewidth]{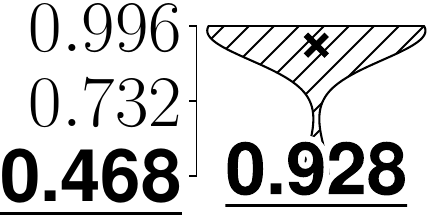}
\end{minipage} & \begin{minipage}{.095\linewidth}
  \includegraphics[width=\linewidth]{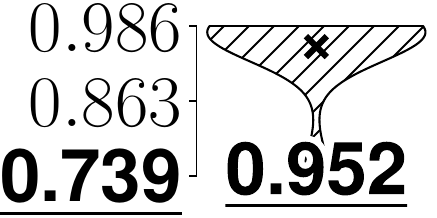}
\end{minipage} & \begin{minipage}{.095\linewidth} 
  \includegraphics[width=\linewidth]{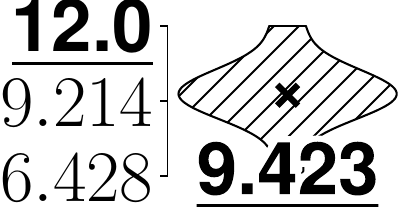}
\end{minipage}\\[2.5mm]\Cline{0.5pt}{2-13}
 \end{tabular}}%
\label{tab:results}
\end{table*}
\begin{table*}
\vspace{-0.4cm}
\raggedright
\scalebox{0.85}{
\begin{tabular}{p{1.1cm}l}
&$^1$ The table includes the graphical distributions of the metrics, and their averages emphasized and marked by \textbf{x}. \\
&$^2$ ``P." denotes \emph{program}. 
\end{tabular} }
\vspace{-0.1cm}
\end{table*}

\section{Application}\label{sec:application}
In this section, we evaluate the performance of various robot programs applied to a machine-tending task. Machine tending involves a robotic system managing tasks like part loading/unloading, machine activation, and monitoring \cite{owen-hill_what_nodate}, aiming to enhance efficiency, productivity, and safety. Fig. \ref{fig:CNC} illustrates the sequential steps involved in the process.

\textbf{\textit{Material Loading:}} The robot identifies and picks an object for processing, considering its shape, position, and orientation, typically using vision systems and sensors. For this study, we use a fixed “Picking Position” for object retrieval. The robot then loads the object into the machine. 

\textit{\textbf{Machine Operation:}} In this mockup, machine operation is represented as idle time, with the robot remaining static. The idle duration varies across experiments to assess the robot’s performance during these periods.

\textit{\textbf{Material Unloading:}} After processing, the robot unloads the object at a fixed “Unloading Position”, avoiding collisions with the environment.

\vspace{-2mm}
\subsection{Materials}
\begin{figure}[t!] 
\centering
\includegraphics[width=.94\linewidth]{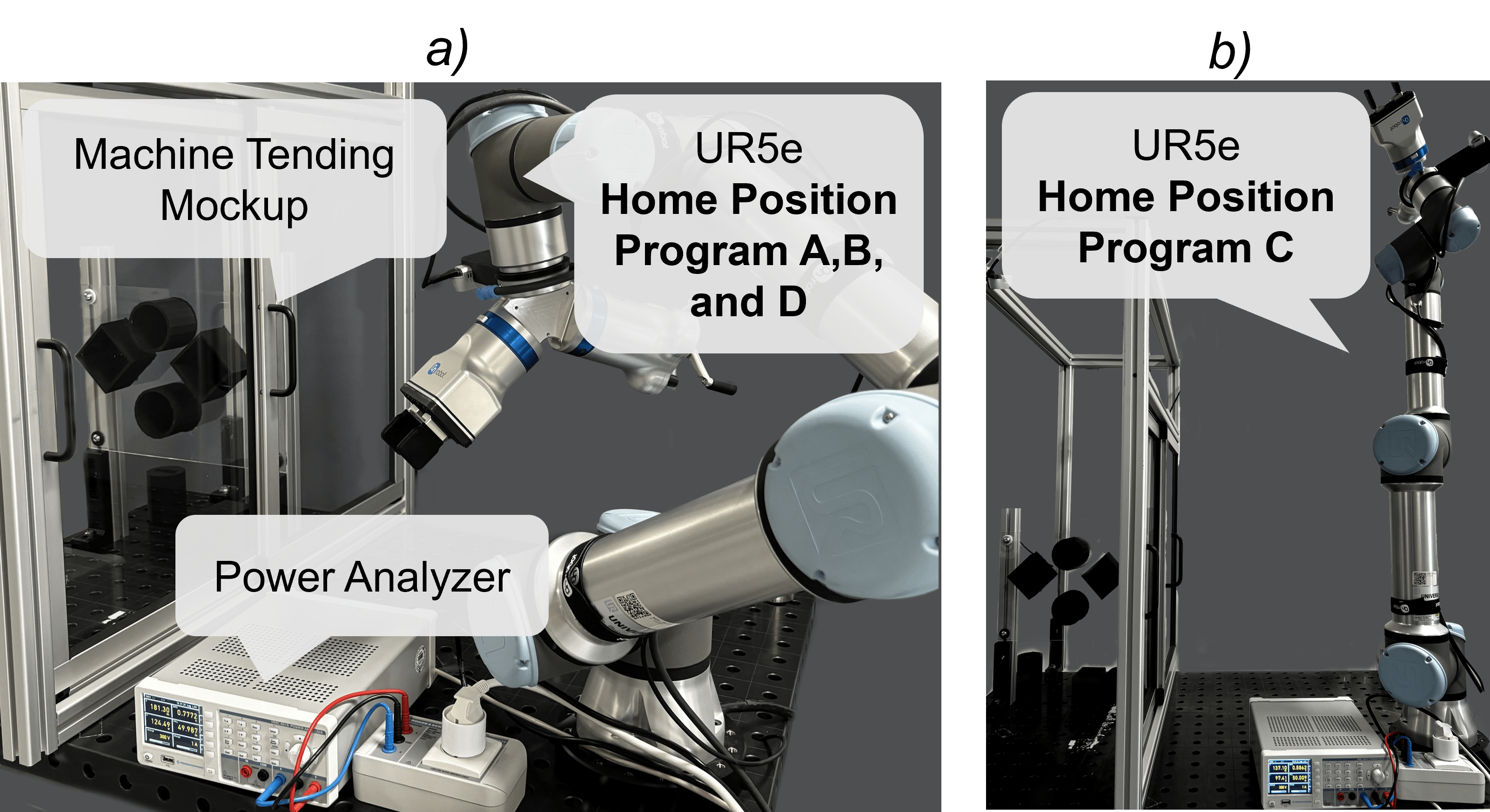}
\vspace{-3mm}
\caption{Test Setup composed of a machine tending mockup, a UR5e, two gripers 2FG7 and 3FG25, and a power analyzer HMC8015. }
\label{fig:setup}   
\vspace{-7mm} 
\end{figure}

\vspace{-1mm}
The experimental setup, shown in Fig. \ref{fig:setup}, includes:

\begin{description}[style=unboxed, leftmargin=0.3cm]
    \item [\textit{Test Cell:}] A simplified Machine Tending setup with a sliding manual door and a chuck with 50 mm radius cylinder holders.
    \item[\textit{Objects:}] Metal cylinders with a 45 mm radius and 2.5 kg weight.
    \item[\textit{Manipulator:}] A UR5e robot, a six-degree-of-freedom manipulator with a 5 kg payload. 
    \item[\textit{Grippers:}] Two OnRobot grippers, 2FG7 (7 kg payload) and 3FG25 (15 kg payload). 
    \item[\textit{Power Analyzer:}] The HMC8015 from Rohde Schwarz, measuring energy consumption with a maximum input of 600 VRMS, 20 A, and a sampling rate of 500k sample/s.  
    \item[\textit{Sensors:}] Velocity and torque measurements from the robot’s internal sensors, accounting for electromagnetic joint torque and friction, however, particular to UR-robots, joint torque estimation is based on motor currents and the motor model. 
\end{description}

\vspace{-3mm}
\subsection{Robot Program}
\vspace{-1mm}

The robot is programmed in two phases: material loading and object unloading, using the manufacturer’s teaching pendant and programming language. We created four distinct programs for the same task, keeping cell positions like “Picking Position,” “Unloading Position,” and “Chuck Position” constant.
The programs differ in picking approach, intermediate waypoints, command types, idle standing position, door handling strategies, and object loading approach to the chuck. The home configuration, where the robot starts idling and ends movement, also vary.

\small
{
\small
\quoteON
    ```
    \textbf{Program A}: Uses Cartesian and angular movements without blending radii. Doors are moved with Cartesian linear movements, and loading is done from the top using force control. The initial idling position is shown in Fig. \ref{fig:setup}.
    ```
\quoteOFF
}

{
\small
\quoteON
    ```
    \textbf{Program B}: Similar to Program A but includes blend radii in movements. Door operations and loading approach are the same. Initial idling position is similar to Program A.
    ```
\quoteOFF
}
{
\small
\quoteON
    ```
    \textbf{Program C}: Like Program B, with blend radii for movements. Door actions and loading approach remain consistent. Initial idling position minimizes torque to reduce energy consumption, as shown in Fig. \ref{fig:setup}.
    ```
\quoteOFF
}
{
\small
\quoteON
    ```
    \textbf{Program D} (Faulty program): This program is similar to Program B but introduces a faulty loading approach that increases the risk of failures. The gripper applies only the minimum force required to grasp the object, making it unreliable and prone to dropping it. Additionally, the force control for placing the object is not fine-tuned, increasing the likelihood of collisions with the machine chuck or improper placement in the holder. These factors reduce the program's reliability and could lead to performance inconsistencies.
    
    ```
\quoteOFF
}
\normalsize

\textbf{General Experimental Conditions} 

\begin{itemize}
\item The initial and final robot configurations are the same, therefore, the potential energy variation ($\Delta E_P$) from Eq. \ref{eq.ebasal} is zero.
\item For each program the experiment is repeated 10 times, such that $n = 10$. 
\item This experiment examined two conditions of machine operation (time between loading and unloading operations): 10 (s) and 100 (s).
\end{itemize}
\vspace{-2mm}

\subsection{Results}
In this section, we present and analyze the experimental outcomes of the machine-tending case study. The performance of four distinct robot programs is evaluated using our energy efficiency and reliability metrics. All measurements are derived from 10 repeated experiments per program and are discussed in the context of the theoretical formulations presented in Sections \ref{sec:ee_metrics} and \ref{sec:r_metrics}.

\subsubsection{\textbf{Energy Performance}} Tab. \ref{tab:results}  summarizes the key energy efficiency metrics for the evaluated programs. The main components include:

\begin{description}[style=unboxed, leftmargin=0.5cm, labelindent=0cm, itemindent=0pt, labelsep=0.3cm]
\item[\textit{Basal Energy Consumption ($E_B$)}] is calculated using Eq. \ref{eq.ebasal}. The electrical power of the electronic components ($P_E$) and the mechanical brakes ($P_{MB}$) are measured using the protocol presented in \cite{heredia_ecdp_2023}. The results show that $P_E$ is 91.14 (W) and $P_{MB}$ is 7.12 (W). 
\item[\textit{Total Robot Energy Consumption ($E_R$)}] is directly measured using the HMC8015 power analyzer, capturing the energy used by the entire robotic system (manipulator and gripper) over the task duration.
\item[\textit{Energy Utilization Efficiency ($f_U$)}]  is estimated based on Eq. \ref{eq.fU}.
\item[\textit{Positive Mechanical Energy ($E_{MP}$)}] is based on Eq. \ref{eq.EM}, relies on the robot's internal sensors, i.e., velocity and torque sensors. 
\item[\textit{Energy Conversion Coefficient ($f_C$)}] is calculated based on Eq. \ref{eq.fC}.
\end{description}




Under a 10-second idling period, Program A exhibits the highest robot energy consumption $E_R$ at 9939$J$, while Program C shows the lowest at 9341$J$. Notably, Program D achieves the best $f_U$ at 0.853. However, paired t-tests indicate no statistically significant differences in $f_U$ among the programs. Across all programs, the energy conversion coefficient reveals that less than 2.5\% of the consumed energy is effectively transformed into useful mechanical work

Under a 100-second idling period, Program A again shows the highest energy consumption $E_R$ at 18645$J$, and Program C the least at 18195$J$.  In this scenario, Program C achieves a significantly higher $f_U$ (0.919), with statistical significance difference (p $<$ 0.05) confirmed by a paired t-test compared to the rest of programs. Notably, the lower idling energy consumption of Program C is attributed to its optimized idling position, which reduces the electrical power draw by approximately 5 W, resulting in reduced energy use and a superior utilization coefficient during extended idling. The conversion coefficients $f_C$ across all programs remain low (less than 2\%), confirming that only a small fraction of consumed energy is converted into mechanical work.

\begin{figure}[t!] 
\centering
\vspace{2mm}
\includegraphics[width=.85\linewidth]{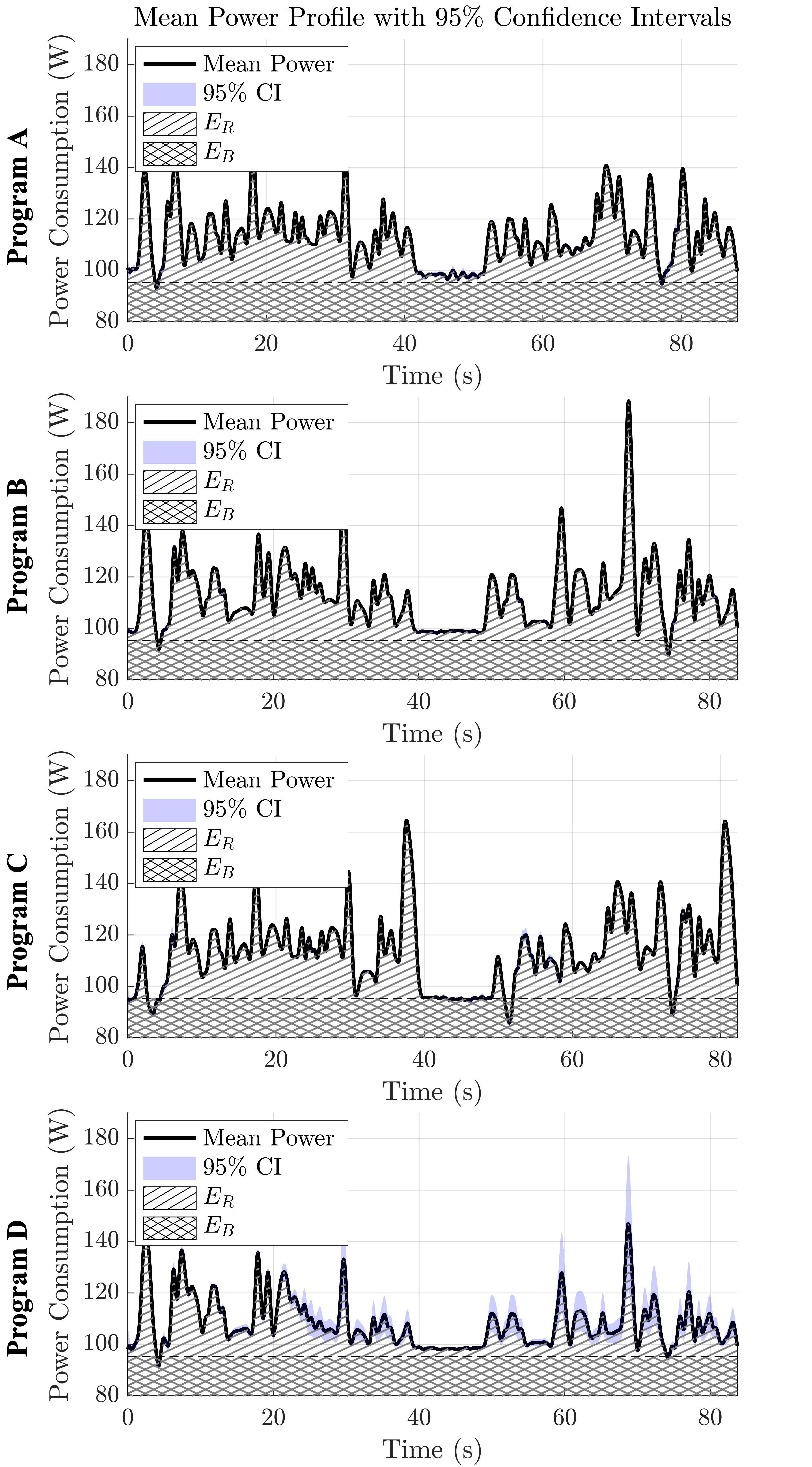}
\vspace{-.3cm}
\caption{This figure illustrates the power profiles of each program including the 95 \% Confidence Interval (CI) for a waiting time of 10 (s). }
\vspace{-.4cm}
\label{Fig:ResultProfiles}    

\end{figure}

\subsubsection{\textbf{Reliability Performance}} Tab. \ref{tab:results} also presents the results of the following reliability metrics:

\begin{description}[style=unboxed, leftmargin=0.5cm, labelindent=0cm, itemindent=0pt, labelsep=0.3cm]
    \item [\textit{Reliability Coefficient $f_R$}] is based on the coefficients $c_1$, $c_2$, and $c_3$, according to Eq. \ref{eq.fR}. The weights on Eq.\ref{eq.fR} are equal for each factor, such that ($w_1 = 1$, $w_2 = 1$, and $w_3 = 1$).
    \item[\textit{Stress metric ($\alpha$)}] is determined using Eq. \ref{eq:eval_alpha}.
\end{description} 

 \begin{figure}[t!]
\centering
\vspace{2mm}
\includegraphics[width=.84\linewidth]{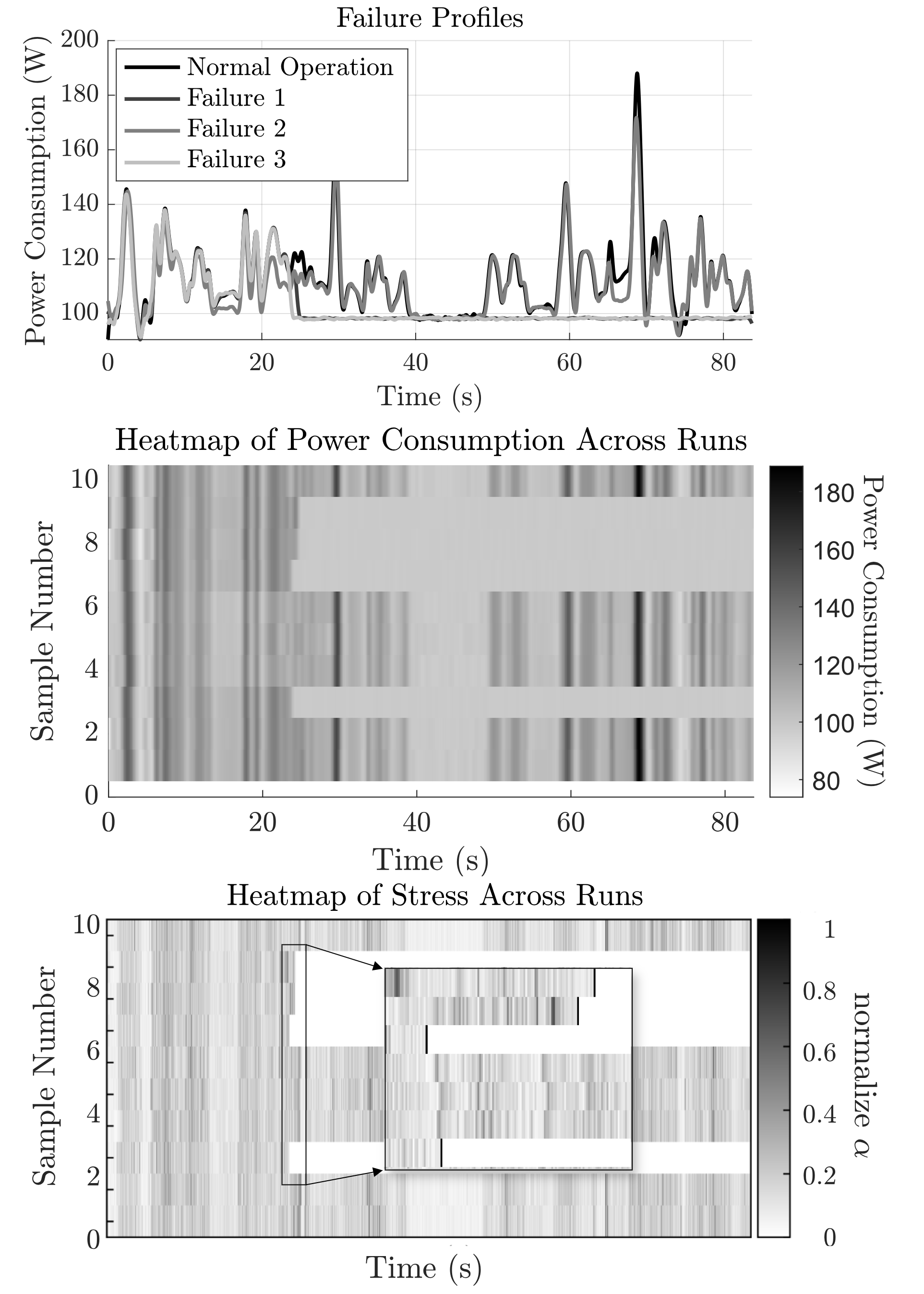}
\vspace{-.5cm}
\caption{This figure illustrates the performance of Program D for 10 (s). Notably, the robot's power profiles show significant changes when the program encounters failures. In Failure 1 (samples 4,5 and 6), the robot fails to pick up the object. In Failures 2 (samples 8 and 9) and 3 (samples 3 and 7), the robot collides with the machine chuck and machine tending mockup, respectively. The power consumption and stress metric for each sample and time are presented utilizing heat maps.}
\label{Fig:ProgramD}
\vspace{-.5cm}

\end{figure}

Statistical analysis reveals that the faulty Program D has the worst reliability, with $f_R$ values of 0.673 and 0.739 for 10- and 100-second idling times, respectively. These differences are statistically significant (p $<$ 0.05). In contrast, the other programs show average $f_R$ values above 0.977, with minimum values not dropping below 0.974. 

In our stress analysis, Program B outperforms Programs A, C, and D in terms of the robot wear metric ($\alpha$), recording values of 8.477 for a 10-second idle time and 9.284 for a 100-second idle time. Program A, which employs Cartesian linear movements, generally induces higher joint wear due to the increased mechanical load associated with these movements. Although Program C demonstrates better energy efficiency owing to its optimized idling position, the additional movement required to reach this position results in increased mechanical stress. Conversely, the failures observed in Program D generate significant stress, particularly during abrupt braking movements where the robot comes to a sudden stop. These findings suggest that while energy efficiency improvements are valuable, they must be balanced against the potential for increased mechanical wear and the risk of failure-induced stress.

Fig. \ref{Fig:ResultProfiles} shows the energy profiles of the robotic system over operation time, including the 95\% confidence interval (CI) for 10 seconds. High reliability systems have overlapping profile curves with small deviations. For program D, the CI area increases due to power differences from failures.

Fig. \ref{Fig:ProgramD} provides further insight into the reliability performance of Program D under a 10-second idling period. The figure includes three sub-figures: the first displays the power profile, highlighting three distinct type of failures; the second sub-figure presents a heat map of power consumption that visually captures the temporal fluctuations and variability across the ten experimental repetitions; and the third sub-figure shows a stress heat map, revealing increased stress levels preceding the failure events, particularly during runs 2, 6, 7, and 8, which are likely attributable to collision impacts or the activation of protective stop mechanisms in the UR5e controller. These visualizations strongly correlate with our theoretical reliability metrics. The observed spikes in power consumption and stress immediately preceding failure events validate our hypothesis that abrupt fluctuations in these signals are indicative of unreliable performance. Moreover, when compared to more reliable programs, the distinct variability in Program D’s profiles supports the use of these metrics as predictive indicators of performance degradation. This detailed analysis not only deepens our understanding of Program D's deficiencies but also highlights the potential for real-time monitoring to inform maintenance strategies and improve overall system reliability.

\begin{figure}[t!] 
    \centering
    \vspace{2mm}    \includegraphics[width=.95\linewidth]{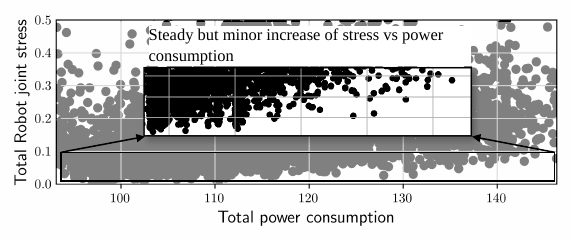}
    \vspace{-.5cm}
    \caption{This figure illustrates a heatmap representation of robot stress across multiple runs in Program A. It emphasizes on the sixth run heatmap, which is derived from power consumption data for comparative analysis.}
    \label{fig:tracking}
    \vspace{-.6cm}
\end{figure}

Examining the relationship between robot wear metric $\alpha$ and power consumption reveals that higher power usage does not necessarily correspond to increased stress levels. Instead, the stress–power correlation exhibits a steady yet minor positive correlation, as illustrated in Fig. \ref{fig:tracking}. Physically demanding tasks, such as object lifting or rapid movement, cause only slight stress increases.  In contrast, pronounced stress spikes are primarily associated with abrupt state transitions (e.g., collisions and safety brakes' activation). By jointly analyzing power consumption and the wear metric, we can track variations in task execution and object handling over time, offering a more comprehensive assessment of the robot’s operational performance.

\vspace{-2mm}
\section{Discussion}\label{sec:discussion}

Our method outlines a protocol to compare robot program performance based on reliability and energy efficiency, using operational measurements during robot operation. This dual-faceted assessment is sensitive to both the specific program and the robot executing it therefore it analyzes the robot from an embodiment perspective .

Energy efficiency metrics offer a standardized perspective on robot performance. Initially designed for comparing programs executing the same task,  these metrics are versatile enough to assess energy efficiency across a variety of tasks. The energy utilization metric $f_U$ relies on external power measurements, making it applicable to a wide range of robots and programs since the power analyzer can be connected to any system without requiring changes in the setup. On the other hand, the energy conversion metric $f_C$ depends on internal sensor data, which means that comparing $f_C$ across robots from different manufacturers requires additional calibration because internal sensors are fixed and not industrially standardized.

Similarly, our reliability metrics are based on the premise that consistently executed tasks should yield consistent energy consumption. Deviations in power profiles serve as indicators of performance variations. However, subtle performance changes may be masked by the complex interplay of electrical and mechanical factors. To improve the discriminative power of the reliability coefficient ($f_R$), its weighting parameters should be calibrated according to the application context. Machine learning methods can support the automatic optimization of these parameters. Since our method relies on external power analyzer measurements, it can be readily extended across different robotic platforms without requiring changes to the robot’s internal architecture.

In our machine tending case, where the robot loads and unloads a cylindrical object, the proposed metrics effectively identify programs that excel in both energy efficiency and reliability. Program C demonstrates the highest energy efficiency during extended idling periods, whereas Program D exhibits the poorest reliability. Our methodology for assessing reliability and energy efficiency can be extended to other applications, offering a valuable framework for identifying best practices in robot programming.

In real industrial applications, achieving a balance between reliability and energy consumption is crucial. Program B exemplifies this balance, outperforming Programs A and D in both energy efficiency and reliability, though it remains less energy efficient than Program C. However, Program B places less mechanical stress on the robot than Program C due to differences in their home configurations. These findings highlight that while optimizing for energy efficiency is important, it must be carefully weighed against mechanical stress and reliability to ensure long-term operational performance.


\vspace{-1mm}
\section{Conclusions}\label{sec:conclusion}
\vspace{-3mm}

This article presents a method for evaluating robot program performance from an embodiment perspective, utilizing the power profile of robot operation. By introducing metrics such as energy utilization $f_U$, energy conversion efficiency $f_C$, the reliability coefficient $f_R$, and the established robot wear metric $\alpha$, we provide a framework that highlights the trade-offs between energy consumption and reliable performance. We validated this method in a machine-tending scenario using a UR5e manipulator, assessing four different programs. The results show that energy efficiency and reliability are not always aligned, emphasizing the need to balance energy consumption with mechanical stress and performance consistency. While Program C demonstrated superior energy efficiency, it also induced higher mechanical stress, whereas Program B achieved a more favorable trade-off. The proposed method can be applied across various robotic applications, contributing to the development of more sustainable and reliable automation solutions.

\vspace{-2mm}
\bibliographystyle{IEEEtran}
\bibliography{references,references2}

\end{document}